\documentclass[10pt, a4paper]{article}

\usepackage[final]{lrec-coling2024}

\usepackage{natbib}
\usepackage{makecell}
\usepackage{multibib}
\usepackage{longtable}
\usepackage{CJK}
\usepackage{multirow}
\usepackage{longtable}
\usepackage{array}
\usepackage{multirow}
\usepackage{array}
\usepackage{graphicx}

\title{KIT-19: A Comprehensive Korean Instruction Toolkit on 19 Tasks for Fine-Tuning Korean Large Language Models}

\name{Dongjun Jang, Sungjoo Byun, Hyemi Jo, Hyopil Shin} 

\address{Department of Linguistics, Seoul National University \\
         3-327, 1, Gwanak-ro, Gwanak-gu, Seoul, Republic of Korea \\
         \{qwer4107, byunsj, huimei6361, hpshin\}@snu.ac.kr\\}

\abstract{
Instruction Tuning on Large Language Models is an essential process for model to function well and achieve high performance in specific tasks. Accordingly, in mainstream languages such as English, instruction-based datasets are being constructed and made publicly available. In the case of Korean, publicly available models and datasets all rely on using the output of ChatGPT or translating datasets built in English. In this paper, We introduce \textit{KIT-19} as an instruction dataset for the development of LLM in Korean. \textit{KIT-19} is a dataset created in an instruction format, comprising 19 existing open-source datasets for Korean NLP tasks. In this paper, we train a Korean Pretrained LLM using \textit{KIT-19} to demonstrate its effectiveness. The experimental results show that the model trained on \textit{KIT-19} significantly outperforms existing Korean LLMs. Based on the its quality and empirical results, this paper proposes that \textit{KIT-19} has the potential to make a substantial contribution to the future improvement of Korean LLMs' performance.
 \\ \newline 
 }

\begin{document}

\maketitleabstract

\section{Introduction}
Pretrained LLM (Large Language Models, \citet{shanahan2022talking, brown2020language, taylor2022galactica, chowdhery2022palm, touvron2023llama} are Transformer-based language models \citep{vaswani2017attention} with an extensive parameter count, typically in the hundreds of billions or beyond, and they undergo training on vast corpora. These models are typically employed for their intended purposes following instruction tuning methodology, a technique aimed at optimizing models to effectively adhere to user-provided instructions. 


Since the release of InstructGPT \citep{ouyang2022training}, instruction datasets in English have been widely developed and made publicly available across various domains such as \citet{ouyang2022training, taori2023alpaca}. However, for Korean, which falls into the category of a relatively low-resource language, there is a notable absence of datasets created in the native language. Notably, the Korean instruction datasets currently available are either created by translating existing English datasets using DeepL\footnote{\url{https://www.deepl.com/translator}} or by relying on outputs from large language models like the ChatGPT API\footnote{\url{https://openai.com/blog/chatgpt}}, without adequately capturing the cultural nuances of the Korean language (Table \ref{tab:kor_instrcution}). Hence, to enhance and advance the performance of Korean Large Language Models, it is imperative to have datasets constructed specifically in the Korean language.

We introduce \textit{KIT-19} (A Comprehensive Korean Instruction Toolkit on 19 Tasks), a universal dataset for Korean Instruction Tuning. \textit{KIT-19} is a dataset constructed for Instruction Tuning, derived from 19 different NLP datasets in the Korean language, each consisting of 5,000 examples. It follows the methodology of \citet{longpre2023flan, bach2022promptsource} which integrates existing NLP datasets into an Instruction Dataset, without relying on machine-translated outputs from other languages or utilizing LLM output as a training dataset. \textit{KIT-19}, as compared to the datasets currently employed for Korean LLM modeling, effectively captures the cultural features of the Korean language and, due to the comprehensive nature of its 19 distinct datasets, contributes to the model's generalizability.

In this paper, we transparently disclose the construction process of \textit{KIT-19} and provide detailed information about the source of the 19 datasets, respectively. To assess the quality of the \textit{KIT-19} dataset, we conduct Full Fine-tuning of our dataset on Polyglot-Ko-5.8b and Polyglot-Ko-1.3b \citet{polyglotko}, which are the Korean representative Pretrained LLMs.

Finally, we evaluate Korean Large Language Models publicly available on a total of 6 benchmark datasets for assessment. The results show that the performance of the Polyglot-Ko-5.8b model trained with \textit{KIT-19} outperforms others, and we could also observe that the Polyglot-Ko-1.3b model exhibits higher performance compared to other Korean LLMs.

The contributions of our study are as follows:
\begin{itemize}
\item We construct and release 100K Korean instruction datasets, addressing the data scarcity problem and reducing the reliance on translated or GPT-generated instructions for Korean LLMs.
\item We demonstrate the efficacy of our \textit{KIT-19} by evaluating LLMs using various benchmarks, comparing those trained with \textit{KIT-19} and those without.

\end{itemize}

\begin{table*}[ht!]
    \centering
    \begin{tabular}{p{0.3\linewidth}|p{0.55\linewidth}}
        \Xhline{3\arrayrulewidth}
        \textbf{Dataset}  & \textbf{Construction Method}\\ \Xhline{3\arrayrulewidth}
        \textbf{KoAlpaca v1.0\footnote{\url{https://huggingface.co/datasets/Bingsu/ko_alpaca_data}}} & After \textbf{translating} the Alpaca instruction, the output is generated by \textbf{ChatGPT}.   \\ \hline
        \textbf{KoAlpaca v1.1\footnote{\url{https://raw.githubusercontent.com/Beomi/KoAlpaca/main/KoAlpaca_v1.1.jsonl}}} & 
After collecting questions from JiSikIn\footnote{\url{https://kin.naver.com/}} (Korean online knowledge-sharing platform), answers were generated using \textbf{ChatGPT}.
  \\ \hline
        \textbf{sharegpt\_deepl\_ko\footnote{\url{https://huggingface.co/datasets/junelee/sharegpt_deepl_ko}}}  &   \textbf{Translated} the ShareGPT data using DeepL.\\ \hline
        \textbf{ShareGPT-74k-ko\footnote{\url{https://huggingface.co/datasets/dbdu/ShareGPT-74k-ko}}} & \textbf{Translated} the cleaned version of ShareGPT with 90k using Google Translate. \\ \hline
         \textbf{KoChatGPT\footnote{\url{https://github.com/airobotlab/KoChatGPT}}} &    After collecting questions from the Korean question dataset, answers were generated using \textbf{ChatGPT}.\\ \hline
          \textbf{OIG-small-chip2-ko\footnote{\url{https://huggingface.co/datasets/heegyu/OIG-small-chip2-ko}}}  & \textbf{Translated} the OIG-smallchip-2\footnote{\url{https://github.com/LAION-AI/Open-Instruction-Generalist}} English data from LAION AI using \textbf{Google Translate}.  \\ \hline
           \textbf{Korquad-Chat\footnote{\url{https://huggingface.co/datasets/heegyu/korquad-chat-v1}}} &Given the context (paragraphs from news and Wikipedia) of the KorQuAD v1 data, relevant conversations were generated using \textbf{ChatGPT}.   \\ \hline
            \textbf{AIRC-KETI/kowow\footnote{\url{https://github.com/AIRC-KETI/kowow}}}  &
The \textbf{translated} data of WoW (Wizard of Wikipedia) - a knowledge-based conversation dataset.  \\ \hline
             \textbf{CounselGPT\footnote{\url{https://github.com/MrBananaHuman/CounselGPT}}} &  Counseling data generated by \textbf{ChatGPT API}.  \\ \hline
              \textbf{Evolve-instruct\footnote{\url{https://github.com/lcw99/evolve-instruct/}}}   &  Instructions were augmented using evol-instruct from WizardLM, and then answers were generated by \textbf{ChatGPT}.\\ \hline
               \textbf{KULLM v2\footnote{\url{https://huggingface.co/datasets/nlpai-lab/kullm-v2}}} & \textbf{Translated} the GPT4ALL, Dolly, and Vicuna (ShareGPT) data using DeepL.   \\ \hline
         \textbf{namuwiki\_alpaca\_dataset\footnote{\url{https://huggingface.co/datasets/psymon/namuwiki_alpaca_dataset}}} & A dataset modified to fit Stanford Alpaca training from the NamuWiki (Korean wiki platform) dump file.    \\ \hline
             \textbf{ko-lima-vicuna\footnote{\url{https://huggingface.co/datasets/changpt/ko-lima-vicuna}}}   & A dataset regenerated in Korean using the \textbf{ChatGPT API}, based on the lima\_vicuna\_format\footnote{\url{https://huggingface.co/datasets/64bits/lima_vicuna_format}} data. \\ \hline
              \textbf{ko-lima\footnote{\url{https://huggingface.co/datasets/taeshahn/ko-lima}}} & A dataset \textbf{translated} into Korean from the training data of \citet{lima} \\ \hline
               \textbf{Ko-StrategyQA\footnote{\url{https://huggingface.co/datasets/NomaDamas/Ko-StrategyQA}}}  & This dataset is the Korean version of StrategyQA\footnote{\url{https://allenai.org/data/strategyqa}}. All questions and paragraphs from the original dataset were \textbf{translated} using DeepL. \\ \hline
             \textbf{KOpen-platypus\footnote{\url{https://huggingface.co/datasets/kyujinpy/KOpen-platypus}}}  &  \textbf{Translated} \citet{platypus} with DeepL.\\ \hline
        \textbf{EverythingLM-data-V2-Ko\footnote{\url{https://huggingface.co/datasets/ziozzang/EverythingLM-data-V2-Ko}}} & \textbf{Translated} EverythingLM V2\footnote{\url{https://huggingface.co/datasets/totally-not-an-llm/EverythingLM-data-V2}} in DeepL.  \\ \hline
        \textbf{human-rights-corpus/HRC/}  & Using the precedents and consultation cases from the National Human Rights Commission of South Korea as a reference, \textbf{GPT-3.5-turbo} for one-shot learning was employed to generate question-answer pairs. \\ 
    \Xhline{3\arrayrulewidth}
    \end{tabular}
    \caption{Existing Korean Instruction Datasets and Construction Methods: It is problematic that most of the existing instruction datasets rely on translation and ChatGPT.}
    \label{tab:kor_instrcution}
\end{table*}

\begin{figure*}[ht]
  \centering
\includegraphics[width=0.95\linewidth]{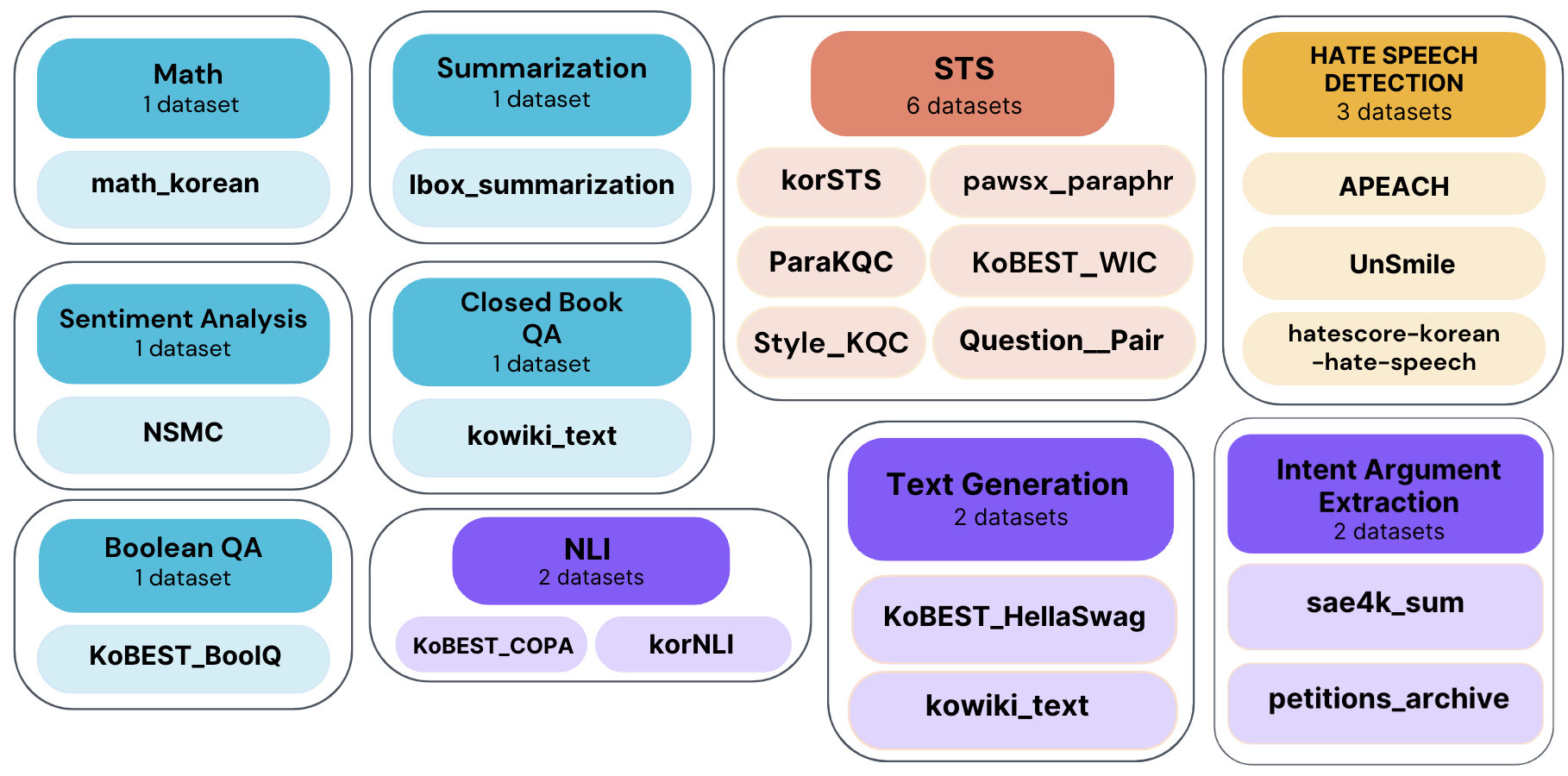}
  \caption{A glance at the KIT for Korean LLM: We create instruction datasets by drawing from 19 Korean NLP datasets across 10 different categories. We utilize `kowiki\_text' as a source dataset for both Closed Book QA and Next Sentence Prediction tasks.}
  \label{fig:data}
\end{figure*}

\section{Related Work}
Instruction tuning has emerged as a potent method when pre-trained language models are further trained to comply with specific user-defined instructions. \citet{instruction} presented a mixture of over 60 NLP datasets expressed via natural language instructions. \citet{alpaca} released 52K instruction data used to fine-tune the Alpaca model. Additionally, \citet{natural_lang_instruction} encapsulated instructions constructed by researchers to crowdsource pre-existing NLP datasets, encompassing 61 unique NLP tasks and a total of 193k instances. The dataset presented by \citet{p3} is the outcome of integrating 170 English NLP datasets and 2,052 English prompts for the purpose of instruction fine-tuning. Other noteworthy open-source instruction datasets in English include UnifiedQA \citep{unifiedqa}, LIMA \citep{lima}, and those generated using Chat-GPT4 \citep{peng}.

While multilingual instructions, exemplified by \citet{super_natural_lang_instruction}, are in existence, open-source instruction datasets for languages other than English remain limited in number. Focusing on Korean, the Koalpaca dataset\footnote{\url{https://github.com/Beomi/KoAlpaca}} adopts translations from DeepL to render Alpaca instructions in the Korean language. The instruction set used for KULLM training \citet{kullm} draws its foundation from English datasets, specifically \citet{peng}, \citet{vicuna}, and \citet{dolly}, translating them into Korean. \citet{kolima} follows a similar path, presenting a Korean translation of the LIMA dataset \citet{lima}. Datasets like Counsel GPT\footnote{\url{https://github.com/MrBananaHuman/CounselGPT}} and Human Rights Corpus (HRC) \citep{right} are crafted using Chat-GPT\footnote{\url{https://openai.com/blog/chatgpt}} and are original Korean instructions, not relying on translations. However, these are specialized datasets and include fewer than 1.5K instances each.

\section{KIT-19 (A Comprehensive Korean Instruction Toolkit on 19 Tasks)}\label{sec:dataset}
Most Korean Instruction datasets rely on translating from English datasets or using the output of high-performing models like ChatGPT. Table \ref{tab:kor_instrcution} lists the currently available open-source Korean instructions and provides their construction methods\footnote{\url{https://github.com/HeegyuKim/open-korean-instructions}}. All the Korean instruction datasets mentioned above, except for the namuwiki\_alpaca\_dataset, rely on machine translation or the ChatGPT generation. Furthermore, each dataset lacks a comprehensive set of tasks, which means that there's a potential for the model to become biased toward specific tasks during the Instruction Tuning process. While it may be acceptable for a model designed solely for one task, for creating models capable of performing a wide range of comprehensive tasks, a comprehensive instruction dataset is required.

Accordingly, in our research, we introduce \textit{KIT-19}\footnote{\url{https://huggingface.co/datasets/snunlp/KIT-19-ToolKit-100000}} (A Comprehensive Korean Instruction Toolkit on 19 Tasks), which integrates the 19 existing open-source NLP datasets as the instruction dataset in Korean. We collected 5000 instances for each of the 19 tasks, summing up to a total of $100K$ instances. Notably, one of these 19 datasets, kowiki\_text is seperated in two distinct categories, which is why the overall instance count is $100K$. Following the approach proposed by \citet{instruction} and \citet{T5}, we collected each dataset within \textit{KIT-19} with a ratio of 5,000 instances per task to ensure that the dataset does not exhibit bias toward specific tasks. Thus, we restrict the number of instances from each dataset to 5,000. Figure \ref{fig:data} displays the source NLP datasets and the categories within our extensive instruction dataset.

\begin{figure*}[t]
  \centering
\includegraphics[width=1.0\linewidth]{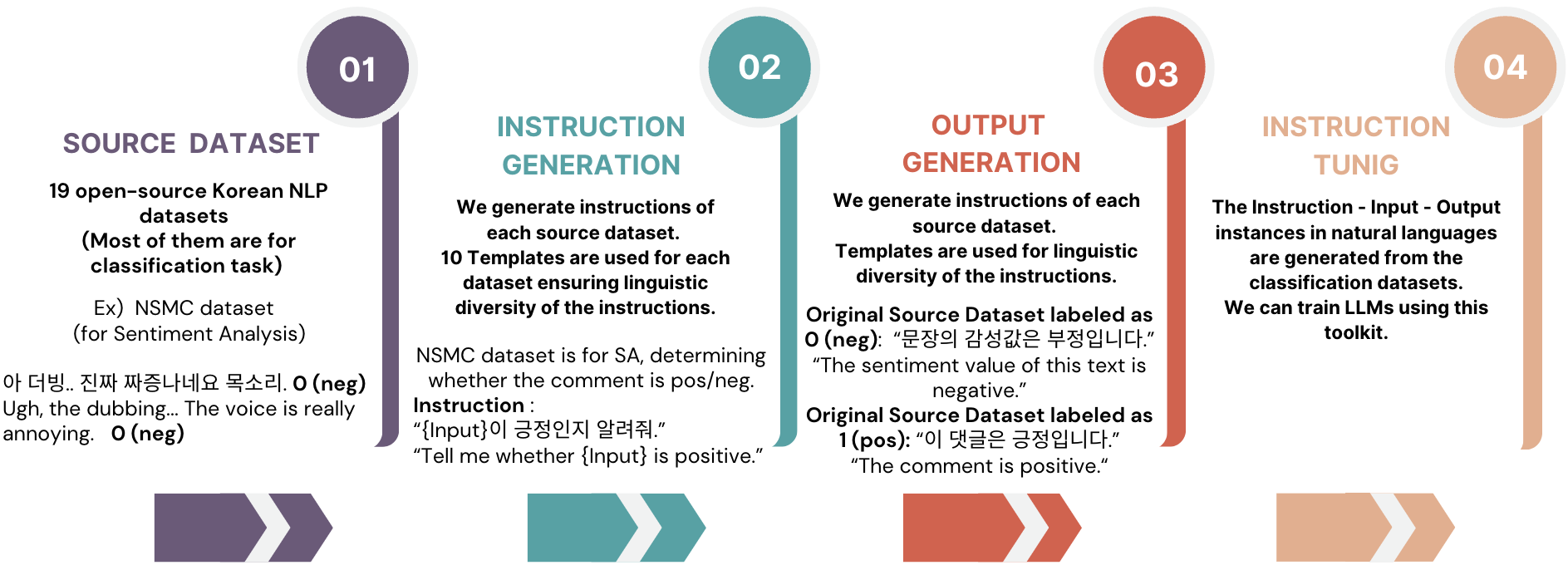}
  \caption{Overview of the data construction procedure of \textit{KIT-19}}
  \label{fig:process}
\end{figure*}

The format of \textit{KIT-19} follows that of \citet{alpaca}, including \textit{Instruction, Input,} and \textit{Output}. Instruction describes the task the model should perform. Input is an optional context or input for the task. For example, when the instruction is `Tell me the definition of the term', the terminology should be the input. Output is an required response that we expect the model to generate. Figure \ref{fig:process} shows the construction procedure of \textit{KIT-19}.

We categorize the 19 tasks comprising \textit{KIT-19} into 10 top-level tasks and introduce the methodology for transforming the objectives of each dataset into an Instruction Dataset (three primary components: Instruction, Input, and Output). In this section, we hope that our explanations would be beneficial for future works in constructing Instruction datasets.

\subsection{Hate Speech Detection}
Hate Speech Detection is the task of determining whether an input sentence contains hate speech or derogatory expressions. Relying solely on translated versions of this task is ineffective, as hate speech is deeply rooted in a language's unique cultural, political, and linguistic nuances. It is also noteworthy that closed generative models such as ChatGPT and CLOVA X \footnote{\url{https://clova-x.naver.com/welcome}} do not process or produce unethical content, thereby hindering the generation of instruction datasets for this task. We generate hate speech detection instruction datasets from three existing NLP datasets: APEACH \citep{apeach}, UnSmile \citep{unsmile}, and hatescore \citep{hatescore}. 

\begin{itemize}
\item \textbf{Instruction}: The instructions primarily focus on questions pertaining to hate speech detection. However, the design of these instructions varies depending on the specific source dataset. Depending on the unique attributes of the original data (for example, whether the dataset categorizes content simply as containing hate speech or not, or if it annotates the target of the hate speech), the instructions might be framed as `Does this \{Input\} contain offensive content?' or `Who or what is the target of this hate speech?'.

\item \textbf{Input}: The input for this task corresponds to the online comments provided by APEACH, UnSmile, and hatescore datasets. 

\item \textbf{Output}: The output is tailored based on the characteristics of each source dataset. For instance, if the original dataset categorizes content in a binary manner, determining whether it contains hate speech, the output is phrased as `The comment is hate speech.'. On the other hand, if the source dataset provides annotations on the target of the hate speech, the output is framed as `The text contains derogatory language aimed at LGBT.'.
\end{itemize}

\subsection{Boolean Question Answering (QA)}
Boolean QA requires the model to respond with either a `Yes' or `No' based on the provided context. We use the KoBEST\_BoolQ \citep{kobest} to generate instructions, input, and output.

\begin{itemize}
\item \textbf{Instruction}: The instruction is typically phrased as a question, such as `Is the question or the statement in the {Input} True or False?'

\item \textbf{Input}: For this task, we use paragraphs and corresponding questions from the KoBEST\_BoolQ dataset. Example is as follows:

\textit{Paragraph}: The term `French Revolution' mainly refers to the 1789 revolution, although it technically encompasses both the July Revolution of 1830 and the February Revolution of 1848. The 1789 event often stands out and is termed the `Great French Revolution.'

\textit{Question}: Isn't the July Revolution part of the Great French Revolution?

\item \textbf{Output}: The output provides a clear True or False response based on the context in the paragraph. For instance, the response might be, `Based on the paragraph, the answer to your question is False.'
\end{itemize}

\subsection{Natural Language Inference (NLI)}
In the NLI task, it is essential to discern the semantic relationship between a given `hypothesis' and `premise' to determine if they entail, contradict, or are neutral with respect to each other. We derive our NLI instructions from the KoBEST\_COPA \citep{kobest} and korNLI \citep{kornli} datasets. Outputs based on KoBEST\_COPA categorize the relationship as cause and effect between two sentences. In contrast, those based on korNLI indicate if the sentences are in entailment, contradiction, or neutral relation. The guiding question for this task is `What is the relationship between the two provided sentences?'.

\begin{itemize}
\item \textbf{Instruction}: The questions mainly address the semantic relationship between two input sentences. For instance, `How are two sentences, \{Input\}, semantically related?'.

\item \textbf{Input}: The paired sentences from the KoBEST\_COPA and korNLI datasets are the input for this task.

\item \textbf{Output}: Outputs are designed according to the specifics of each source dataset. For example, using the KoBEST\_COPA dataset, the output might state `Sent 1 is the cause and Sent 2 is the effect.', whereas with the korNLI dataset, it might say `These sentences contradict each other.'
\end{itemize}

\subsection{Text Generation}
Text Generation involves creating a sentence or a paragraph that logically follows an input text. We transform two existing datasets into instruction sets for this Text Generation task: KoBEST\_HellaSwag \citep{kobest} and kowiki\_text \footnote{\url{https://github.com/lovit/kowikitext}}. For KoBEST\_HellaSwag, the output is a sentence that naturally succeeds the input text. In contrast, for kowiki\_text, the output is a full paragraph. The kowiki\_text dataset is primarily based on the Korean Wikipedia. For data generation, we split the text in half, using the first part as input and the latter half as output. 

\begin{itemize}
\item \textbf{Instruction}: The primary prompt for this task is `Produce text that naturally continues from the presented text.'

\item \textbf{Input}: For KoBEST\_HellaSwag, the preceding sentence from the original dataset serves as the input. Meanwhile, for kowiki\_text, we use the first section of the divided Wikipedia text.

\item \textbf{Output}: The output is the succeeding sentence or paragraph drawn from the source datasets.
\end{itemize}

\subsection{Semantic Textual Similarity (STS)}
Semantic Textual Similarity (STS) is a task that measures the semantic similarity between two sentences or texts. For this task, we source from 6 datasets: korSTS \citep{kornli}, pawsx\_paraphr \citep{paws-x}, ParaKQC \citep{parakqc}, KoBEST\_WIC \citep{kobest}, Style\_KQC \citep{style}, and Question\_Pair \footnote{\url{https://github.com/songys/Question_pair}}.

\begin{itemize}
\item \textbf{Instruction}: A typical instruction for the STS task might ask, `Do these sentences convey the same meaning?'

\item \textbf{Input}: We utilize pairs of sentences or texts from the previously mentioned six datasets as our input.

\item \textbf{Output}: The output evaluates how similar in meaning the input sentences are. A possible response could be, `Yes, the sentences are semantically alike.' However, for the korSTS dataset, which provides a quantified similarity score ranging from 0 to 5, an output might be phrased as, `The similarity between the two provided texts scores 4.2 out of 5.'
\end{itemize}

\begin{figure*}[t]
  \centering
\includegraphics[width=0.8\linewidth]{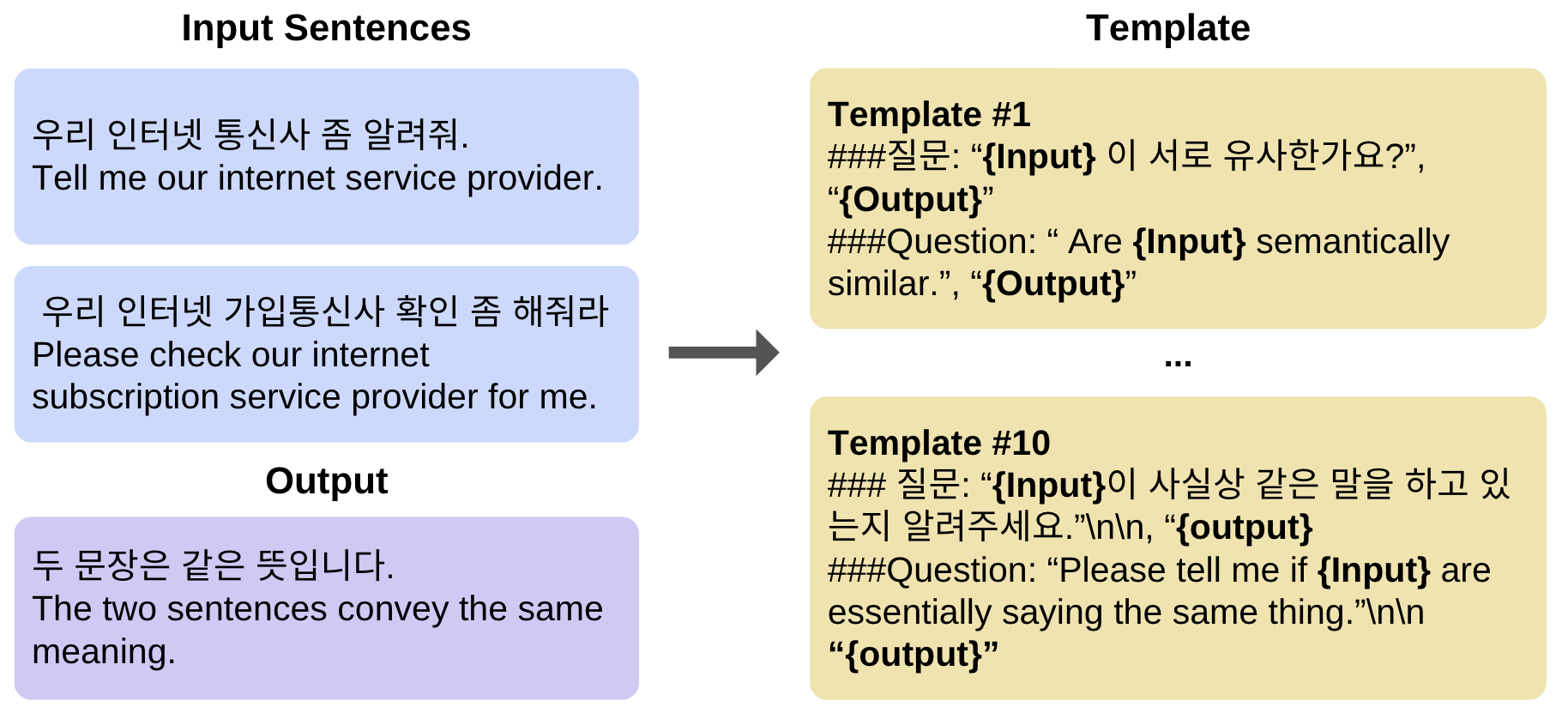}
  \caption{Instruction Template used to construct \textit{KIT-19}. Above is the example of one of the datasets used for STS task. We employ 10 unique templates for each dataset, resulting in a total of 200 templates.}
  \label{fig:template}
\end{figure*}

\subsection{Sentiment Analysis (SA)}
Sentiment Analysis (SA) determines if a text has a positive or negative tone. Generating instruction datasets for this task is challenging if machine translation or ChatGPT are used since it is difficult to process euphemisms, typos, or profanities commonly found in online comments. The Naver Sentiment Movie Corpus, namely NSMC dataset \citep{NSMC} consists of movie comments\footnote{\url{http://movie.naver.com/movie/point/af/list.nhn}}, which are labeled either 0 for negative or 1 for positive sentiments. Rather than translating English SA instruction datasets or generating instructions with ChatGPT, we use NSMC datasets to generate three components as follows.:
\begin{itemize}
\item \textbf{Instruction}: This is a textual string, asking to determine the sentiment value of the input sentence, such as `Is the provided text positive or negative?'

\item \textbf{Input}: Our input consists of movie reviews sourced from the NSMC dataset.

\item \textbf{Output}: Instead of binary responses like 0 or 1, we phrase the outputs in natural language, such as `The provided text has a negative tone.'
\end{itemize}

\subsection{Intent Argument Extraction}
Intent Argument Extraction is aimed at discerning the primary purpose or intent embedded within a text. We leverage two NLP datasets for guiding this category: sae4k\_sum \citet{sae4k_sum} and petitions\_archive \footnote{\url{https://github.com/lovit/petitions_archive}}. sae4k\_sum comprises \textit{question--command--intent} pairs and the labels for classification of the utterance type. petitions\_archive provides data sourced from public petitions presented to the Blue House (Cheong Wa Dae). This dataset offers both the content of these petitions and their respective titles. We interpret the title as encapsulating the core intent of the petition. As such, when crafting outputs, we utilize the title segment.

\begin{itemize}
\item \textbf{Instruction}: For this particular task, the guiding question is: `What is the author's intent behind the text?'

\item \textbf{Input}: We draw from the question/command portion of the sae4k\_sum dataset and the main text from the petitions\_archive dataset for our inputs.

\item \textbf{Output}: We use the intent segment from the sae4k\_sum dataset as the target output. For the petitions\_archive dataset, its title serves as the output to indicate intent.
\end{itemize}

\subsection{Math}
This task requires the model to provide accurate answers to mathematical problems. We create mathematical instructions using the math\_korean dataset\footnote{\url{https://github.com/jkc-ai/mwp-korean-data-2021}}.

The original dataset is in JSON format and includes key information such as class (the type of mathematical question, e.g., Size comparison or Arithmetic operations), question (the math problem itself), answer, and an equation. We transform this JSON dataset into three components: instruction, input, and output.
\begin{itemize}
\item \textbf{Instruction}: This is a textual string, for example, `What is the correct answer to this math problem?' 

\item \textbf{Input}: This corresponds to the question from the original dataset. For instance, `Students in Shinhye's class surveyed their favorite sports among the students. Among those surveyed, the ratio of students who like swimming and soccer is (2/5) of those who like swimming, and the students who like soccer are twice as many as those who like swimming. If there are 10 students who like swimming, how many students like soccer?'

\item \textbf{Output}: The output is structured as follows: `The answer is 8, as calculated by 10 * (2/5) * 2.' Both the answer and the equation from the original JSON dataset are utilized in generating the output.
\end{itemize}


{\renewcommand{\arraystretch}{1.3} 
\begin{table*}[t]
\centering
\resizebox{\textwidth}{!}{%
\begin{tabular}{l|l|c|c|c|c|c|c}
\Xhline{3\arrayrulewidth}
& \textbf{Metric} &{\textbf{Polyglot-ko-1.3b}} & {\textbf{Polyglot-ko-5.8b}} & {\textbf{KoAlpaca-5.8B}}  & {\textbf{kullm-polyglot-5.8b-v2}} & {\textbf{KIT-5.8b}} & {\textbf{KIT-1.3b}}\\ \Xhline{3\arrayrulewidth}
\multirow{2}{*}{\textbf{KoBEST\_COPA}} & ACC & 72.00\%  & 77.60\%  & 69.80\%  & 76.60\% & \textbf{91.60\%}  & 83.80\%  \\
& F1 (macro) & 71.96\%  & 77.55\%  & 69.77\%  &76.53\% & \textbf{91.59\%}  & 83.78\%  \\
\hline
\multirow{2}{*}{\textbf{KoBEST\_BoolQ}} & ACC & 49.86\% & 53.63\%  & 56.34\%   &50.28\%  & \textbf{66.24\%}  & 50.71\%  \\
& F1 (macro) & 35.52\%  & 43.56\% & 50.64\% & 33.71\%   & \textbf{66.14\%}  & 34.78\% \\
\hline
\multirow{3}{*}{\textbf{KoBEST\_HellaSwag}} & ACC & 40.60\%  & 48.80\%  & 38.20\%  & 44.40\% & \textbf{97.60\%} & 81.60\% \\
& ACC\_Norm & 53.00\%  & 59.80\%  & 46.20\% &55.20\%  & \textbf{98.20\%}  & 89.80\% \\
& F1 (macro) & 40.13\%  & 48.53\% & 38.15\%  &44.25\%& \textbf{97.61\%}  & 81.49\% \\
\hline
\multirow{2}{*}{\textbf{KLUE\_STS}} & ACC & 42.39\%  & 45.28\%  & 51.83\% &42.39\% & \textbf{65.51}\% & 42.20\% \\
& F1 & 59.54\%  & 60.34\%  & 33.86\%  &59.54\% & \textbf{69.71\%}  & 56.52\% \\
\hline
\multirow{2}{*}{\textbf{KoBEST\_SentiNeg}} & ACC & 69.27\%  & 50.38\%  & 38.79\%  & 50.38\%  & 71.54\%  & \textbf{80.86\%}\\
& F1 & 68.19\% & 33.95\%  & 38.48\%  & 33.50\% & 68.98\%  & \textbf{80.86\%}\\
\hline
{\textbf{KLUE\_YNAT}}
& F1 & 33.24\% & 33.62\%  & 20.91\%  & 32.20\% & 28.15\%  & \textbf{38.34\%} \\
 \Xhline{3\arrayrulewidth}
\end{tabular}%
}
\caption{Experiment Results: We evaluate 7 models on 6 different benchmark datasets. KIT-5.8b and KIT-1.3b, which use our own templates as shown in Figure \ref{fig:template}, out-perform the others. Original templates were used for the other models. Detailed experimental setups and additional results can be found in Appendix \ref{sec:appendixA}}
\label{tab:experiment}
\end{table*}
}

\subsection{Closed Book QA}
Closed Book QA involves a model answering open-domain questions without relying on external knowledge. In this task, we create instructions using the kowiki\_text dataset\footnote{\url{https://github.com/lovit/kowikitext}}. While this dataset has previously been used in Text Generation tasks, we approach it differently in this task. Instead of splitting the Wikipedia text in half, we use the definitions or descriptions of terminology. The dataset is transformed into three components as follows.

\begin{itemize}
\item \textbf{Instruction}: This consists of a textual string, such as `Provide information about \{Input\}.'

\item \textbf{Input}: The input for this task corresponds to the title of the wiki text provided by kowiki\_text. In other words, it represents the main topic or subject being questioned in the instruction. For instance, `Ivan Aivazovsky's life' could be an example of input data.

\item \textbf{Output}: The output is structured based on the content found in kowiki\_text. For example, `Ivan Aivazovsky was born in a small village in Feodosiya, on the Crimean Peninsula of the Russian Empire, to a poor Armenian family. His parents' last name was Aivazian...'
\end{itemize}

\subsection{Summarization}
Summarization task requires a model to summarize the key content of the input text. We use lbox\_summarization dataset \citep{lbox} to generate instructions. lbox\_summarization dataset consists of 20K \textit{precendent -- summary} pairs. We use 5K pairs among them to convert the dataset into instructions, input, and output. 

\begin{itemize}
\item \textbf{Instruction}: This consists of a textual string, such as `Briefly summarize \{Input\}.'

\item \textbf{Input}: The input for this task corresponds to the main precedent of the original lbox\_summarization dataset.

\item \textbf{Output}: The output is structured based on the summary part of the lbox\_summarization dataset.
\end{itemize}

\section{200 Instruction Templates for the Utilization of \textit{KIT-19} During the Training Process} 
\textit{KIT-19} is a Comprehensive Dataset consisting of 10 categories. To fully capture the characteristics of each task, we introduced the process of instructifying each task in Section \ref{sec:dataset}. In Section \ref{sec:dataset}, we have organized 10 templates for each task, enabling the utilization of the constructed datasets for their respective objectives. These templates differ in aspects like sentence structure, the use of honorifics, and phrasing. We craft templates based on the specific features of each source dataset to maintain its distinct features. Furthermore, we diversify the output of each task to ensure that the model can provide a variety of responses. We generate outputs considering the characteristics of the source datasets.

\section{Experimental Setup}
We believe that \textit{KIT-19} is a well human-crafted dataset as the Korean instruction dataset. When evaluating the quality of the constructed dataset, both the dataset construction process and distribution are essential, but the performance enhancement when applying the dataset to a model is also a critical factor. In this section, we assess the performance of KIT-5.8b and KIT-1.3b models trained on Polyglot-Ko-5.8b and Polyglot-Ko-1.3b models, respectively, with the utilization of \textit{KIT-19}.

For evaluation, we follow the method of \citet{eval-harness}. We used the polyglot branch of lm-evaluation-harness to utilize Korean benchmark datasets implemented for evaluation. Models are evaluated on two types of templates: an original template provided by lm-evaluation-harness and a custom template used in training KIT models as shown in Figure 3. By varying the format of evaluation templates, we examine the extent to which the evaluation score is affected by the accordance of templates used in training and evaluation. We set $batch\_size$ as 32 to facilitate comparison with other Korean LLMs and $num\_fewshot$ as 0, 1, 2, 3 for the ablation study.

\subsection{Models and Benchmarks}\label{sec:benchmark}
As baseline models, we test Polyglot-Ko-1.3b, Polyglot-Ko-5.8b, Koalpaca-5.8b\footnote{\url{https://github.com/Beomi/KoAlpaca}}, and KULLM \citep{kullm} models which are the mostly employed in the Korean NLP fields. 

For benchmarks, we use KoBEST\_COPA, BoolQ, HellaSwag, SentiNeg \citep{kobest}, KLUE\_STS, and KLUE\_YNAT \citep{klue} to investigate whether the instruction tuning improved the model’s performance both on seen tasks and unseen tasks. Among these, SentiNeg, KLUE\_STS, and KLUE\_YNAT (KLUE-TC) are unseen datasets that are not included in \textit{KIT-19}. We choose (macro) f1 as our evaluation metric to match the guideline presented in the literature and to facilitate comparison with models including polyglot. Every KOBEST benchmark and KLUE\_STS present F1 as their proper evaluation metric except KLUE\_YNAT which chooses macro F1. (footnote: Note that KLUE\_STS differs from the other two unseen tasks in that one of its source corpora, PARAKQC, overlaps with our KIT-19 dataset, which makes KLUE\_STS a partially unseen task.) We select macro F1 for KOBEST and F1 for KLUE among the evaluation metrics available in the evaluation framework to follow the literature as much as possible.

\subsection {Hyper-Parameters}
We train the models for 1 epoch, using a batch size of 8 for the KIT-1.3b model and 4 for the KIT-5.8b model, employing 1024 gradient accumulation steps. We set the learning rate at 3e-5. We use four pieces of H-100 GPUs for training process.

\section{Results}

For the evaluation of \textit{KIT-19}, we select Korean LLMs based on the Polyglot-Ko models. Polyglot-Ko represents pretrained LLMs without undergoing Instruction Tuning. The Kullm-polyglot-5.8-v2 model and KoAlpaca-5.8B model are the instruction tuned models that have undergone Instruction Tuning, each utilizing the Polyglot-Ko-5.8b model as a base, as illustrated in Figure \ref{fig:data}. In the case of benchmark datasets, to ensure fairness, we include three of the Unseen Benchmark tasks out of the six tasks, as detailed in Section \ref{sec:benchmark}. Table \ref{tab:experiment} displays the result of the evaluation, showing that the models trained with our KIT-19 outperform in all the benchmark, with $num_fewshot$ of the models’ highest performance varying in each task.

\subsection{The Effectiveness of the \textit{KIT-19} Dataset in Terms of Performance}
As shown in Table \ref{tab:experiment}, it is evident that the KIT-5.8b\footnote{\url{https://huggingface.co/snunlp/KIT-5.8b/}} and KIT-1.3b\footnote{\url{https://huggingface.co/snunlp/KIT-1.3b/}} models significantly outperform existing Korean LLMs. When compared to the other four models, it is clear that the KIT models exhibit the highest performance across all six benchmark sets used in the experiments. Notably, the performance gap compared to other models averages at 15\% higher, underscoring the data quality of \textit{KIT-19} in terms of performance. This also serves as direct evidence highlighting the limitations of Instruction Datasets relying on translation and ChatGPT.

The KIT-5.8b model exhibited the highest performance in four tasks (three Seen datasets and one Unseen dataset), notably excelling in the three Seen datasets of the KoBEST series compared to other models. While this performance improvement could be attributed to the inclusion of the dataset in the training process, our interpretation is that the task-specific ratios within the \textit{KIT-19} dataset played a crucial role in achieving well-rounded performance enhancements without bias toward particular tasks. The model achieved accuracy of over 90\% in the COPA and HellaSwag benchmarks, underscoring the quality and significance of the \textit{KIT-19} dataset.

Finally, the KIT-1.3b model, despite having model parameters over four times smaller than the 5.8b model, outperformed both the KoAlpaca-5.8b and Kullm-Polyglot-5.8b-v2 models. In particular, the KIT-1.3b model exhibited the highest performance in the SentiNeg and YNAT benchmarks. Furthermore, it achieved an accuracy of over 80\% in the COPA and HellaSwag benchmark datasets, thus validating the efficacy of the \textit{KIT-19} dataset.

\subsection{The Necessity for Expanding the Domains of the \textit{KIT-19} Dataset}
\textit{KIT-19} has convincingly demonstrated its effectiveness by outperforming other models in the model performance evaluation process. Moreover, the KIT models exhibited superior performance in the Unseen Benchmarks compared to other models, suggesting that \textit{KIT-19} indirectly facilitated learning in the Out-Domain during model training. However, to ensure stable performance in tasks beyond benchmarks, it is essential to include a variety of domains. In our future research, we plan to expand the \textit{KIT-19} to overcome challenges related to Out-Domain scenarios.

\section{Conclusion}

In this paper, we have addressed the need for native language instruction datasets for Korean LLMs. While Pre-trained LLMs could have robustness with further training, especially when fine-tuned with instruction-style datasets, the availability of instruction datasets in Korean has been notably scarce. Accordingly, we introduce \textit{KIT-19}, filling a critical gap in the development of Korean LLMs. This comprehensive dataset is humanly constructed by integrating 19 existing Korean NLP tasks, each comprised of 5,000 examples. Our work contains the constructing procedure of \textit{KIT-19} for future works, explaining the methodology for transforming the objectives of each tasks into an Instruction Dataset. For each task, we provide 10 templates, allowing them to be used adequately for their specific objectives. Additionally, we diversified the output for each task to ensure the model could generate a variety of responses. For validation of our dataset, we fine-tuned \textit{KIT-19} on two Korean Pre-trained LLMs and evaluated their performance on six benchmark datasets. We also compared the results with those of four other different models. As a result, we observed that our models outperformed other Korean models significantly. This not only highlights the limitations of models relying on translated or LLM-generated datasets but also underscores the essential role of \textit{KIT-19}. Our models exhibited superior performance in the Unseen benchmarks, outperforming other models. Our research team interprets this result as a consequence of indirectly learning unseen information during the training process. For future research, we plan to expand \textit{KIT-19} to include more domains, enhancing the generality of the LLMs.

\section{Ethical Statement}
In our research, we employed existing NLP datasets to produce our instruction datasets. It is imperative to highlight that each of these datasets comes with permissions for modification and distribution. We have reviewed and ensured the adherence to the licensing and copyright stipulations associated with each dataset.

Releasing our dataset to the public is intended to promote progress in Korean language model development. We emphasize the value of open research and our commitment to enhancing the shared intellectual resources of the NLP field.

\nocite{*}
\section{Bibliographical References}\label{sec:reference}

\bibliographystyle{lrec-coling2024-natbib}
\bibliography{lrec-coling2024-example}

\appendix

\section{Analysis of Performance Across Different Templates and Few-Shot Settings}\label{sec:appendixA}

The evaluation of the models trained with \textit{KIT-19} across a range of benchmarks provides insightful observations into their performance, particularly concerning the type of evaluation templates and few-shot settings. This section delves into the differential impacts of template types and the number of shots on model performance. These analyses are encapsulated in four critical evaluations, as highlighted in Tables 3 to 6.

\subsection{Performance in Seen Datasets with the Original Template}

Table 3 presents the performance of models evaluated against Seen datasets using the original template. A discernible pattern emerges, indicating that consistency in the template type used during instruction tuning and evaluation positively impacts model performance. Specifically, models trained with our \textit{KIT-19} dataset, KIT-1.3b and KIT-5.8b, demonstrate superiority across all three Seen datasets under this evaluation condition. 
Moreover, as the number of shots (n=shot) increases, a nuanced performance trend is observed, suggesting that the models benefit variably from the additional contextual information provided by increased shots. This finding underscores the adaptability of models trained with \textit{KIT-19} in leveraging varying amounts of provided context to enhance their performance.

\subsection{Performance in Unseen Datasets with the Original Template}

The evaluation of models against Unseen datasets with the original template, detailed in Table 4, sheds light on the models' generalization capabilities. Notably, the KIT-1.3b and KIT-5.8b models, despite being evaluated on tasks not seen during the training phase, maintain their performance edge, particularly under specific few-shot settings. This robust performance across unseen tasks highlights the comprehensive nature of the \textit{KIT-19} dataset in preparing models for a broader range of tasks beyond those explicitly encountered during instruction tuning.

\subsection{Performance in Seen Datasets with the Custom Template}

Table 5 focuses on evaluating models against Seen datasets but pivots to using the custom template. The consistent outperformance of KIT-models underscores the significance of template consistency between training and evaluation phases. The KIT models exhibit remarkable performance improvements, which can be attributed to the tailored instruction delivery mechanism ingrained through the instruction tuning with \textit{KIT-19}. This approach not only enhances model performance on familiar tasks but also exemplifies the models' ability to adhere to varied instructional formats.

\subsection{Performance in Unseen Datasets with the Custom Template}

Finally, Table 6 illustrates the models' performance against Unseen datasets, evaluated with the custom template akin to the training phase’s template structure. Here, the adaptability of the KIT models is once again highlighted, showing a clear performance advantage in most settings. This demonstrates the models' capability to generalize learned instructions to novel tasks effectively, particularly when the evaluation aligns closely with the training template format. The nuanced examination of zero-shot and few-shot settings further reveals the complex interplay between model training and its application versatility in unseen domains.

{\renewcommand{\arraystretch}{2.0}
\begin{table*}[t]
\centering
\resizebox{\textwidth}{!}{%
\begin{tabular}{cc|cccc|cccc|cccc}
\hline
 &  & \multicolumn{4}{c|}{\textbf{KOBEST\_COPA (n=shot)}} & \multicolumn{4}{c|}{\textbf{KOBEST\_BoolQ (n=shot)}} & \multicolumn{4}{c}{\textbf{KOBEST\_HellaSwag   (n=shot)}} \\ \cline{3-14} 
\textbf{Model} & \textbf{Params} & n=0 & n=1 & n=2 & n=3 & n=0 & n=1 & n=2 & n=3 & n=0 & n=1 & n=2 & n=3 \\ \hline
\textbf{Polyglot-ko-1.3b} & 1.3B & 71.96\% & 71.06\% & 71.06\% & 72.96\% & 35.52\% & 51.27\% & 49.55\% & 51.55\% & 40.13\% & 38.67\% & 41.14\% & 40.35\% \\
\textbf{Polyglot-ko-5.8b} & 5.8B & 77.55\% & 75.56\% & 77.26\% & 76.75\% & 43.56\% & 54.06\% & 56.47\% & 56.97\% & 48.53\% & 48.47\% & 48.85\% & 49.88\% \\
\textbf{KoAlpaca-5.8b} & 5.8B & 69.77\% & 68.25\% & 67.67\% & 68.25\% & 50.64\% & 51.64\% & 54.70\% & 53.93\% & 38.15\% & 39.24\% & 38.87\% & 38.44\% \\
\textbf{kullm-polyglot-5.8b-v2} & 5.8B & 76.53\% & 75.87\% & 76.98\% & 75.48\% & 33.71\% & 53.73\% & 52.71\% & 51.72\% & 44.25\% & 46.17\% & 47.15\% & 47.47\% \\
\textbf{KIT-1.3b} & 1.3B & 84.09\% & 83.99\% & 83.99\% & 84.18\% & 33.71\% & 51.27\% & 50.17\% & 48.58\% & 82.11\% & 80.51\% & 78.34\% & 77.91\% \\
\textbf{KIT-5.8b} & 5.8B & \textbf{91.38\%} & \textbf{92.39\%} & \textbf{93.19\%} & \textbf{93.69\%} & \textbf{64.28\%} & \textbf{61.23\%} & \textbf{64.72\%} & \textbf{65.23\%} & \textbf{98.00\%} & \textbf{97.80\%} & \textbf{97.79\%} & \textbf{97.60\%} \\ \hline
\end{tabular}%
}
\caption{Experiment results in Seen datasets with the original template: We evaluate 7 models on 6 different benchmark datasets.}
\end{table*}
}

{\renewcommand{\arraystretch}{2.0}
\begin{table*}[ht]
\centering
\resizebox{\textwidth}{!}{%
\begin{tabular}{cc|cccc|cccc|cccc}
\hline
 &  & \multicolumn{4}{c|}{\textbf{KLUE-STS (n=shot)}} & \multicolumn{4}{c|}{\textbf{KOBEST\_SentiNeg (n=shot)}} & \multicolumn{4}{c}{\textbf{KLUE\_YNAT (n=shot)}} \\ \cline{3-14} 
\textbf{Model} & \textbf{Params} & n=0 & n=1 & n=2 & n=3 & n=0 & n=1 & n=2 & n=3 & n=0 & n=1 & n=2 & n=3 \\ \hline
\textbf{Polyglot-ko-1.3b} & 1.3B & 59.54\% & 44.87\% & 48.08\% & 49.60\% & \textbf{68.19\%} & 54.16\% & 65.32\% & 64.97\% & 33.24\% & 25.02\% & 26.01\% & 28.15\% \\
\textbf{Polyglot-ko-5.8b} & 5.8B & \textbf{60.34\%} & 45.30\% & 39.83\% & 45.92\% & 33.95\% & 66.47\% & 75.41\% & 79.49\% & 33.62\% & 23.67\% & 26.76\% & 29.85\% \\
\textbf{KoAlpaca-5.8b} & 5.8B & 33.86\% & 45.25\% & 46.12\% & 53.38\% & 38.48\% & 61.59\% & 75.75\% & \textbf{84.11\%} & 20.91\% & 19.15\% & 19.25\% & 19.13\% \\
\textbf{kullm-polyglot-5.8b-v2} & 5.8B & 59.54\% & \textbf{49.03\%} & 46.67\% & 52.42\% & 33.50\% & \textbf{74.64\%} & 68.09\% & 67.45\% & 32.20\% & 21.06\% & 26.26\% & 27.28\% \\
\textbf{KIT-1.3b} & 1.3B & 59.78\% & 44.87\% & 39.63\% & 42.42\% & 46.23\% & 65.23\% & 62.14\% & 65.38\% & \textbf{35.46\%} & \textbf{30.14\%} & \textbf{29.80\%} & \textbf{31.69\%} \\
\textbf{KIT-5.8b} & 5.8B & 58.04\% & 44.97\% & \textbf{50.09\%} & \textbf{57.29\%} & 42.63\% & 68.17\% & \textbf{77.68\%} & 78.73\% & 30.89\% & 24.02\% & 26.89\% & 28.42\% \\ \hline
\end{tabular}%
}
\caption{Experiment results in Unseen datasets with the original template}
\end{table*}}

{\renewcommand{\arraystretch}{2.0}
\begin{table*}[ht]
\centering
\resizebox{\textwidth}{!}{%
\begin{tabular}{cc|cccc|cccc|cccc}
\hline
 &  & \multicolumn{4}{c|}{\textbf{KOBEST\_COPA (n=shot)}} & \multicolumn{4}{c|}{\textbf{KOBEST\_BoolQ (n=shot)}} & \multicolumn{4}{c}{\textbf{KOBEST\_HellaSwag   (n=shot)}} \\ \cline{3-14} 
\textbf{Model} & \textbf{Params} & n=0 & n=1 & n=2 & n=3 & n=0 & n=1 & n=2 & n=3 & n=0 & n=1 & n=2 & n=3 \\ \hline
\textbf{Polyglot-ko-1.3b} & 1.3B & 67.54\% & 66.54\% & 67.06\% & 66.54\% & 33.40\% & 40.53\% & 43.31\% & 42.10\% & 38.40\% & 38.94\% & 40.20\% & 39.74\% \\
\textbf{Polyglot-ko-5.8b} & 5.8B & 70.74\% & 69.15\% & 69.30\% & 69.99\% & 55.93\% & 54.17\% & 56.38\% & 54.49\% & 43.02\% & 44.36\% & 45.28\% & 44.71\% \\
\textbf{KoAlpaca-5.8b} & 5.8B & 65.40\% & 66.55\% & 66.42\% & 66.41\% & 39.86\% & 50.11\% & 50.92\% & 53.66\% & 37.91\% & 36.97\% & 36.16\% & 37.14\% \\
\textbf{kullm-polyglot-5.8b-v2} & 5.8B & 70.54\% & 71.85\% & 72.87\% & 71.65\% & 39.84\% & 55.42\% & 54.27\% & 53.18\% & 41.36\% & 42.73\% & 41.85\% & 44.43\% \\
\textbf{KIT-1.3b} & 1.3B & 83.78\% & 82.49\% & 81.68\% & 80.88\% & 34.78\% & 49.83\% & 48.34\% & 46.41\% & 81.49\% & 78.95\% & 76.94\% & 76.35\% \\
\textbf{KIT-5.8b} & 5.8B & \textbf{91.59\%} & \textbf{90.09\%} & \textbf{90.99\%} & \textbf{91.79\%} & \textbf{66.14\%} & \textbf{60.74\%} & \textbf{58.27\%} & \textbf{59.72\%} & \textbf{97.61\%} & \textbf{96.80\%} & \textbf{96.79\%} & \textbf{96.37\%} \\ \hline
\end{tabular}%
}
\caption{Experiment results in Seen datasets with the custom template}
\end{table*}}

{\renewcommand{\arraystretch}{2.0}
\begin{table*}[ht]
\centering
\resizebox{\textwidth}{!}{%
\begin{tabular}{cc|cccc|cccc|cccc}
\hline
\textbf{} & \textbf{} & \multicolumn{4}{c|}{\textbf{KLUE-STS (n=shot)}} & \multicolumn{4}{c|}{\textbf{KOBEST\_SentiNeg   (n=shot)}} & \multicolumn{4}{c}{\textbf{KLUE-TC (n=shot)}} \\ \cline{3-14} 
\textbf{Model} & \textbf{Params} & n=0 & n=1 & n=2 & n=3 & n=0 & n=1 & n=2 & n=3 & n=0 & n=1 & n=2 & n=3 \\ \hline
\textbf{Polyglot-ko-1.3b} & 1.3B & 59.54\% & 44.87\% & 45.30\% & 33.60\% & 34.06\% & 39.37\% & 50.61\% & 53.35\% & 36.86\% & 22.35\% & 19.36\% & 21.23\% \\
\textbf{Polyglot-ko-5.8b} & 5.8B & 57.90\% & 44.87\% & 34.11\% & 35.10\% & 33.50\% & 34.06\% & 36.03\% & 38.31\% & \textbf{45.50\%} & 15.75\% & 21.09\% & 21.38\% \\
\textbf{KoAlpaca-5.8b} & 5.8B & 58.11\% & 44.87\% & 27.68\% & 28.82\% & 33.50\% & 38.90\% & 51.86\% & 55.20\% & 30.64\% & 22.41\% & 25.62\% & 25.13\% \\
\textbf{kullm-polyglot-5.8b-v2} & 5.8B & 59.34\% & 44.35\% & 35.18\% & 37.00\% & 33.50\% & 54.07\% & 49.03\% & 46.94\% & 38.86\% & 18.62\% & 19.88\% & 21.24\% \\
\textbf{KIT-1.3b} & 1.3B & 56.52\% & 44.87\% & 26.70\% & 28.48\% & \textbf{80.86\%} & \textbf{66.33\%} & \textbf{77.55\%} & \textbf{79.58\%} & 38.34\% & \textbf{28.05\%} & \textbf{29.83\%} & \textbf{32.00\%} \\
\textbf{KIT-5.8b} & 5.8B & \textbf{69.71\%} & \textbf{46.51\%} & \textbf{45.52\%} & \textbf{46.29\%} & 68.98\% & 34.50\% & 33.50\% & 33.50\% & 28.15\% & 15.49\% & 15.02\% & 14.20\% \\ \hline
\end{tabular}%
}
\caption{Experiment results in Unseen datasets with the custom template}
\end{table*}}

\end{document}